\documentclass[10pt, a4paper]{article}

\usepackage{lrec-coling2024} 

\usepackage{graphicx}
\usepackage{subcaption}
\usepackage{multicol}
\usepackage{multirow}
\usepackage{url}
\usepackage{amsmath}
\usepackage{amssymb}

\title{Multilingual Turn-taking Prediction Using Voice Activity Projection}

\name{Koji Inoue$^{1}$, Bing'er Jiang$^{2}$, Erik Ekstedt$^{2}$, Tatsuya Kawahara$^{1}$, Gabriel Skantze$^{2}$}

\address{$^{1}$Graduate School of Informatics, Kyoto University, Japan \\
         $^{2}$Division of Speech, Music and Hearing, KTH Royal Institute of Technology, Sweden \\
         inoue.koji.3x@kyoto-u.ac.jp, binger@kth.se, erikekst@kth.se \\
         kawahara@i.kyoto-u.ac.jp, skantze@kth.se \\}

\abstract{
This paper investigates the application of voice activity projection (VAP), a predictive turn-taking model for spoken dialogue, on multilingual data, encompassing English, Mandarin, and Japanese.
The VAP model continuously predicts the upcoming voice activities of participants in dyadic dialogue, leveraging a cross-attention Transformer to capture the dynamic interplay between participants.
The results show that a monolingual VAP model trained on one language does not make good predictions when applied to other languages.
However, a multilingual model, trained on all three languages, demonstrates predictive performance on par with monolingual models across all languages.
Further analyses show that the multilingual model has learned to 
discern the language of the input signal.
We also analyze the sensitivity to pitch, a prosodic cue that is thought to be important for turn-taking.
Finally, we compare two different audio encoders, contrastive predictive coding (CPC) pre-trained on English, with a recent model based on multilingual wav2vec 2.0 (MMS). 
\\ \newline  }

\begin{document}

\maketitleabstract

\renewcommand{\thefootnote}{\fnsymbol{footnote}}
\footnote[0]{This paper has been accepted for presentation at The 2024 Joint International Conference on Computational Linguistics, Language Resources and Evaluation (LREC-COLING 2024) and represents the author's version of the work.}
\renewcommand{\thefootnote}{\arabic{footnote}}

\section{Introduction} \label{sec:intro}

Turn-taking is a fundamental aspect of spoken interaction between humans, and consequently an important function to model in spoken dialogue systems~\cite{skantze2021review}.
In human-human conversations, the transitioning of the conversational floor is smoothly conducted. 
It has been shown across various languages that the transition offset is typically very brief, around 100-500 msec~\cite{stivers2009universals}.
This indicates that humans use various turn-taking signals across multiple modalities, including lexical cues, prosody, gaze, respiratory, and gestures, in order to coordinate~\cite{wlodarczak2016respiratory,kendrick2023turn}.
In addition, given that the listener also needs some time to articulate a response, there is likely a prediction mechanism involved, where the listener predicts that the speaker's utterance is about to end~\cite{garrod2015use,Levinson2015TurnTaking,ishimoto2017interspeech}.

While recent advancements in large language models (LLMs) have made it easier to generate highly sophisticated responses in spoken dialogue systems, turn-taking is still typically handled in a very simplistic manner.
In practical spoken dialogue systems, turn-taking is commonly implemented using a simple silence timeout threshold, typically around 1 second, to indicate the end of a turn. Silence, however, is not a very good indicator, as silences within turns (pauses) are typically longer than silences between turns, in human-human interaction~\cite{heldner2010pauses}.
This means that spoken dialogue systems are often plagued by long response delays or frequent interruptions in pauses.

To address this problem, many proposals have been made for end-of-turn prediction models.
These models consider verbal and non-verbal cues (such as linguistic and prosodic features) of preceding user utterances, in order to predict whether the user is just pausing (a \textit{hold}), or whether the turn is yielded (a \textit{shift}). 
In earlier models, feature engineering was common, but it has now become more popular to input time-series data, such as prosodic features and word vector representations (word embeddings), into neural networks like recurrent neural networks (RNNs)~\cite{skantze2017sigdial,masumura2017}.
More recently, transformer-based models have been proposed, which can take in the raw input text or audio in an end-to-end processing manner~\cite{ekstedt2020turngpt,sakuma2023slt,muromachi2023interspeech,kurata23_interspeech}.

Another limitation of earlier models has been their sole focus on the binary prediction of turn hold vs. shift.
A more comprehensive model of turn-taking should involve more nuanced decisions. 
When taking the turn, it is, for example, necessary to determine the appropriate waiting time before starting to talk~\cite{raux2012,lala2018icmi,sakuma2023slt}.
There is also a difference in predicting backchannels vs. turn-shifts~\cite{lala2017sigdial}.
As stated above, humans are able to not just react to turn-yielding cues, but they can also predict upcoming turn-shifts.
This would clearly also be a desirable property of spoken dialogue systems. 
Furthermore, there is no established and robust method for handling interruptions and overlaps in conversation, which are commonly observed in human-human conversations.
A more dynamic turn-taking prediction model is required to enable spoken dialogue systems to handle turn-taking in a more human-like manner.
Crucial for such models is that they do not make turn-taking decisions at specific events, but that they operate in a continuous fashion. 

Several models have been proposed recently that make more nuanced turn-taking predictions continuously in a time frame manner~\cite{skantze2017sigdial,lala2019icmi}.
Among such continuous models, the voice activity projection (VAP) model is used in this study~\cite{erik2022vap}.
The VAP model uses multi-layer Transformers and predicts the near future voice activities of dialogue participants by processing the raw audio signals from the two speakers in a dyadic dialogue.
Previous work has shown that the VAP model outperforms other models in predicting turn-taking behaviors, including backchannel predictions~\cite{erik2022vap}.
In the latest version of this model, cross-attention Transformer layers are added after the self-attention layer, to model the audio from the two speakers separately, as explained in Section~\ref{sec:vap}.
Recently, the VAP model has been extended for various purposes including backchannel prediction~\cite{liermann2023emnlp}, multi-modal turn-taking prediction~\cite{onishi2023hai}, and its real-time processing~\cite{inoue2024iwsds}.

To our knowledge, the VAP model has so far only been trained and tested for English~\cite{erik2022sigdial}.
Since it only operates on raw audio, it is technically straightforward to apply it to other languages (even ones it was not trained on).
However, it is not clear how much of turn-taking cues are universal.
In this paper, we investigate to what extent a model trained on one language can be transferred to other languages, but also whether it is possible to build a multilingual model.
This would clearly be more desirable than having separate models trained specifically for different languages.
Specifically, we aim to construct a trilingual model for English, Mandarin Chinese, and Japanese. Those languages were partly chosen since they represent three different language families (Germanic, Sino-Tibetan, and Japonic) and therefore should exhibit a certain level of diversity.

Previous research has analyzed the differences in turn-taking behavior among languages.
Typical differences include the timing of turn-taking~\citep{stivers2009universals, dingemanse2022text}.
For example, the turn transition time of Mandarin and Japanese, centering around 0 msec, while English has more overlaps between turns~\citet{dingemanse2022text}.
Furthermore, in the analysis of turn-taking cues, it has been pointed out that the intonation change at the end of preceding utterances is effective regardless of the language \cite{duncan1972some,local1986towards,koiso1998analysis,ward2000prosodic,levow2005turn,gravano2011turn}.
Specifically, Mandarin shows turn-final pitch lowering for all words in both task-oriented and daily conversations, regardless of the original lexical tone \citet{jian2011mandarin, levow2005turn}.
Moreover, there are differences in the use of backchannels (short utterances such as ``yeah'' and ``yes''), which are essential behaviors in turn-taking, and it has been noted that Japanese has the highest frequency, followed by English and then Chinese~\cite{clancy1996conversational}.
In summary, there are both common tendencies and differences in turn-taking behavior among languages, which justifies the importance of the proposed multilingual turn-taking prediction model.

Achieving a multilingual turn-taking prediction model can lead to the realization of a multilingual spoken dialogue system that does not require specifying the input language.
To our knowledge, this is the first attempt to achieve a multilingual turn-taking model based on the cross-attention Transformer.

Based on the above, this study sets the following research questions:
\begin{itemize}
    \item[] \textbf{RQ1}: Can a VAP model trained on one language be directly applied to another language?
    \item[] \textbf{RQ2}: Is it possible to train a single multilingual VAP model that would be on-par with a monolingual model (trained and evaluated on the same language)?
    \item[] \textbf{RQ3}: Has the multilingual model learned to identify the language?
    \item[] \textbf{RQ4}: How important is pitch for the multilingual model?
    \item[] \textbf{RQ5}: What is the effect of the audio encoder on the model's performance?
\end{itemize}

The rest of this paper is organized as follows:
First, in Section~\ref{sec:vap}, we explain the VAP model that serves as the basis for this study.
In Section~\ref{sec:dataset}, we introduce each of three language dialogue datasets used in this study and analyze the differences among the languages.
We conduct experiments to answer the research questions mentioned above in Section~\ref{sec:exp} and conclude the paper in Section~\ref{sec:conclusion}.

\section{Voice Activity Projection (VAP)} \label{sec:vap}

As stated above, the main objective of the VAP model is to predict a future voice activity of two speakers in a dialogue, based on raw audio input.
For this study, we use the public repository of the VAP model\footnote{\url{https://github.com/ErikEkstedt/VoiceActivityProjection}} to make the results reproducible.
Note that parameters of the VAP model in this study are derived from the above original repository.

\subsection{Model Architecture} \label{sec:vap:model}

\begin{figure}[t]
    \includegraphics[width=\linewidth]{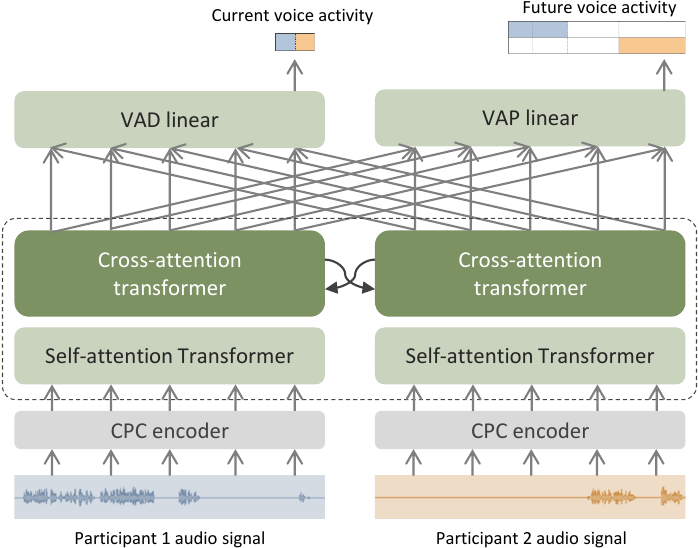}
    \caption{Architecture of the VAP model}
    \label{fig:vap:architecture}
\end{figure}

Figure~\ref{fig:vap:architecture} illustrates the architecture of the VAP model.
The input is a stereo audio signal, with each channel corresponding to each participant's audio.
The length of the input audio signal is assumed to be a maximum of 20 seconds.
In this study, we use a sampling rate of 16 kHz, and a frame rate of 50Hz.

The input signal of each channel is encoded by a pre-trained model of Contrast Predictive Coding (CPC).
The CPC model used in this study is composed of a 5-layer CNN and a single-layer GRU, and it is pre-trained with the Librispeech dataset~\cite{riviere2020unsupervised}.
The unsupervised pre-training algorithm of CPC predicts latent representations of audio in the near future, using a contrastive learning, where negative samples are taken from other time frames.
In this way, the model is similar to the VAP model, in terms of predicting the near future.
Additionally, the pre-trained CPC model has been reported to be useful for multilingual phoneme recognition~\cite{riviere2020unsupervised}, even when English was used for the pre-training.
Thus, we will also rely on a CPC model pre-trained on English. However, in Section~\ref{sec:exp:mms}, we also compare the performance with a multilingual audio encoder (MMS).
During the training of the VAP model, the parameters of the pre-trained CPC are frozen.
The dimension of the vector output by the CPC encoder model is 256.

The vectors encoded by the CPC model are inputted to a self-attention Transformer for each channel.
In this study, we utilize a one-layer Transformer with a dimension of 256.
Subsequently, the outputs from the two channels are fed into a cross-attention Transformer.
In this Transformer, the vector from the first channel serves as the query, while another vector from the second channel acts as the key and value.
The reverse case is also simultaneously performed.
This way, interactive information between the two channels are encoded.
The output is the combined result of these two computations.
This cross-attention mechanism draws inspiration from recent dialogue audio generative models like dGSLM~\cite{nguyen2023generative}.
In our current setup, we employ three-layer Transformers for this interactive mechanism, with a dimension of 256.
The final output of this Transformer is the concatenation of the dual Transformers, resulting in a dimension of 512.
The number of attention heads is set to 4, and the dropout rate during training is set at 0.1.

Finally, the output vector is passed through two separate linear layers for multitask learning.
The first main task is the VAP objective itself, with a dimension of 256 (see next section).
The second task is voice activity detection (VAD), a subtask that detects the current voice activities of the two participants.
The output vector of this subtask has two dimensions, where each dimension corresponds to the voiced probability of each participant.
The future voice activity depends on the current voice activity, so by adding this subtask of VAD, we aim to stabilize the training of VAP.

\subsection{VAP State}

\begin{figure}[t]
    \includegraphics[width=\linewidth]{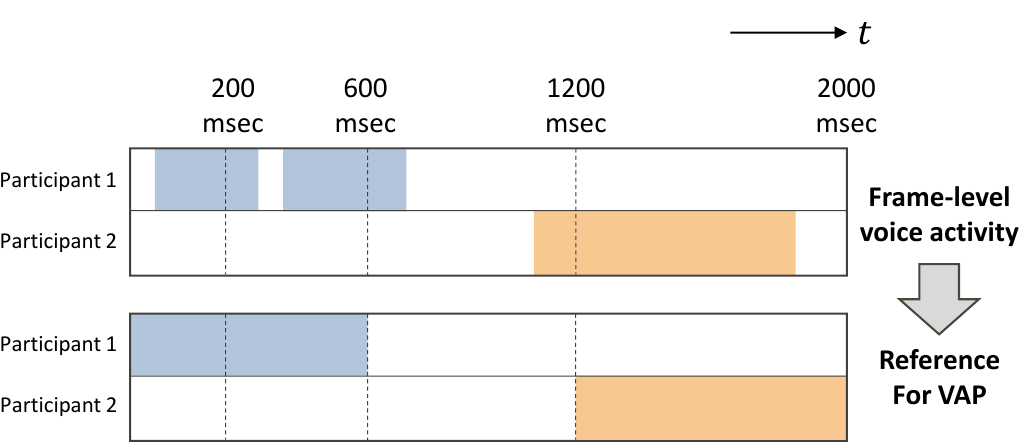}
    \caption{Discretizing bins for the VAP model}
    \label{fig:vap:ref}
\end{figure}

The main objective of the VAP model is to predict the voice activity for both participants within a two-second time window.
Instead of making independent predictions for both speakers (as in~\cite{skantze2017sigdial}), the VAP model makes a prediction of the joint activity of the two speakers over the future time window.
The time window is divided into four binary bins: 0-200 msec, 200-600 msec, 600-1200 msec, and 1200-2000 msec, as depicted in Figure~\ref{fig:vap:ref}.
Since the model assumes two speakers, there are a total of 8 binary bins.
This results in 256 ($=2^8$) combinations of possible activations, representing various events such as turn-shifts, backchannels, overlapping speech, etc.
The objective for the VAP model is formulated as a multi-class classification problem that aims to predict which state out of the 256 patterns the future two seconds will fall into, and the model will output a probability distribution over those states.

Since the ground truth voice activity is often recorded with finer step sizes than those bins, they have to be discretized.
Figure~\ref{fig:vap:ref} illustrates this process, where a bin is defined as ``voiced'' if there are more voiced frames than unvoiced frames in it, and ``unvoiced'' otherwise.

\subsection{Loss Function}

The outputs of the VAP and VAD tasks are applied to the softmax function, and then probabilities of VAP state indexed as $y \in (1, \cdots , 256)$ and voice activity of participant $s$ are calculated as $p_{vap}(y)$ and $p_{vad}(s)$, respectively.
Then, the cross-entropy losses with respect to the reference data are computed as:
\begin{align}
L_{vap} = & - \log p_{vap}(y) ~ , \\
L_{vad} = & - \sum_{s=1}^{2} \{ v_{s} \log p_{vad}(s) \nonumber \\
& + (1-v_{s}) \log (1- p_{vad}(s)) \} ~ 
\end{align}
where $y$ is the index of the reference VAP state, and $v_{s} \in (0, 1)$ is the reference voice activity of participant $s$ (1 for voiced, 0 for unvoiced).
Finally, the losses of both VAP and VAD are combined by adding them together to form the final loss function for optimization as
\begin{equation}
L = L_{vap} + L_{vad} ~ .
\end{equation}
Note that the notation for time frames is omitted due to space limit, although the calculations mentioned above are performed for all input time frames.

\subsection{Turn-taking Prediction Using VAP}

While the probability distribution over the possible VAP states represents a complex prediction of what the turn-taking dynamics will look like in the near future, it can be hard to use and interpret directly.
A simplified representation of the output can be obtained by summing up the probability values of each participant's bins in the 0-200 msec and 200-600 msec regions.
Then, softmax can be applied to both sums to obtain $p_{now}(s)$, which represents a short-term future voice activity prediction of each participant (i.e., ``how likely is each participant to speak in the next 600 msec'').
Similarly, for the 600-1200 msec and 1200-2000 msec bins, $p_{future}(s)$ is used as a slightly longer-term future voice activity prediction.
It is important to note that this is just one example of how the VAP output can be utilized.

\begin{figure*}[t]
    \begin{subfigure}{0.32\textwidth}
    \centering
        \includegraphics[width=\linewidth]{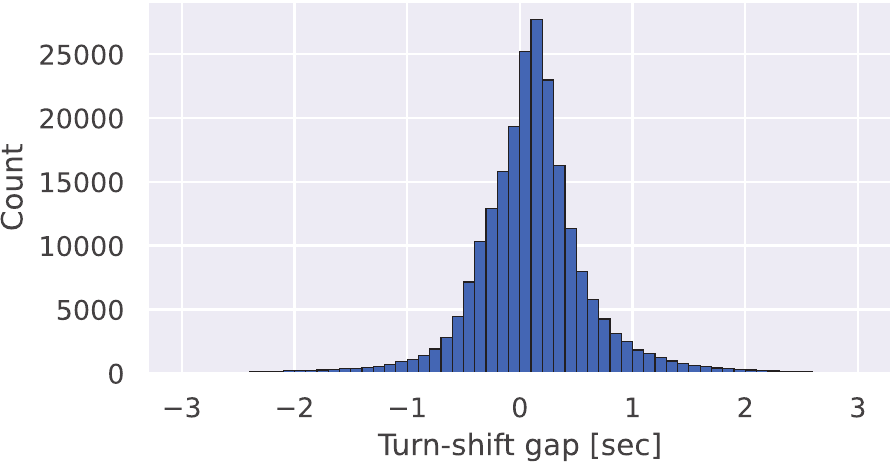}
        \caption{English (Switchboard)}
    \end{subfigure}
    \hfill
    \begin{subfigure}{0.32\textwidth}
    \centering
        \includegraphics[width=\linewidth]{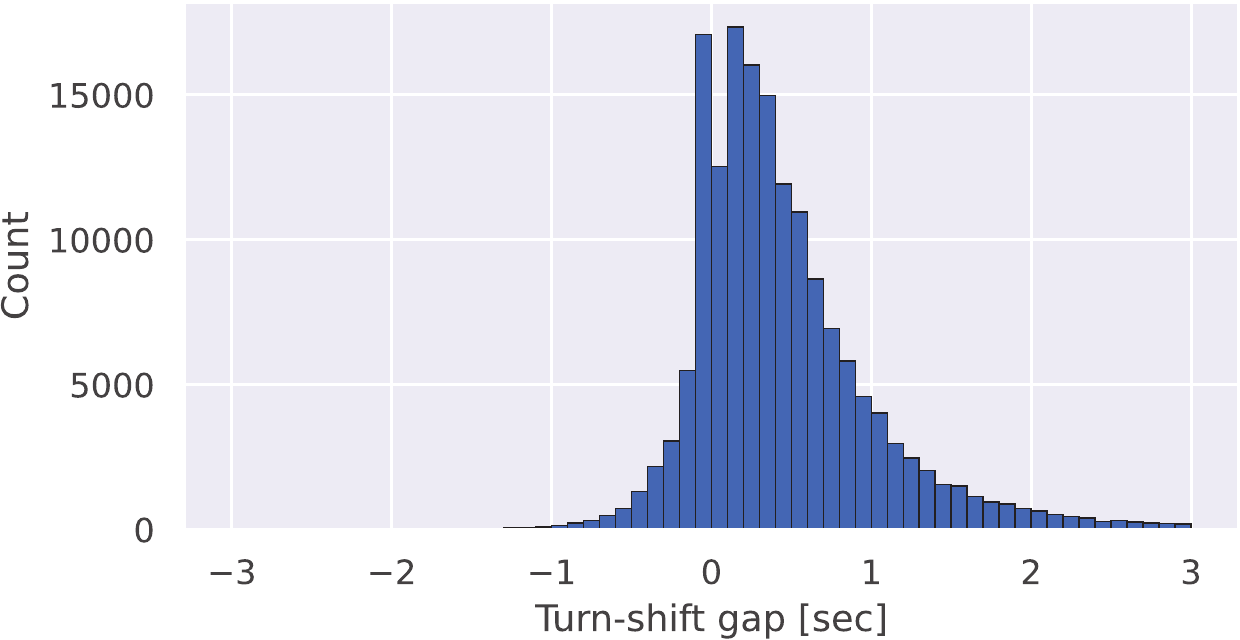}
        \caption{Mandarin (HKUST)}
    \end{subfigure}
    \hfill
    \begin{subfigure}{0.32\textwidth}
    \centering
        \includegraphics[width=\linewidth]{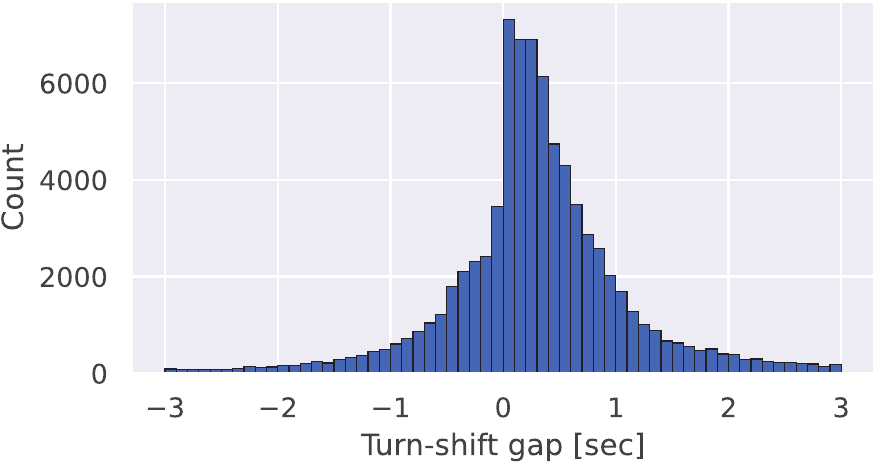}
        \caption{Japanese (Travel agency)}
    \end{subfigure}
    \caption{Histogram of \textbf{turn-shift} gap in three languages}
    \label{fig:hist:gap}
\end{figure*}

\begin{figure*}[t]
    \begin{subfigure}{0.32\textwidth}
    \centering
        \includegraphics[width=\linewidth]{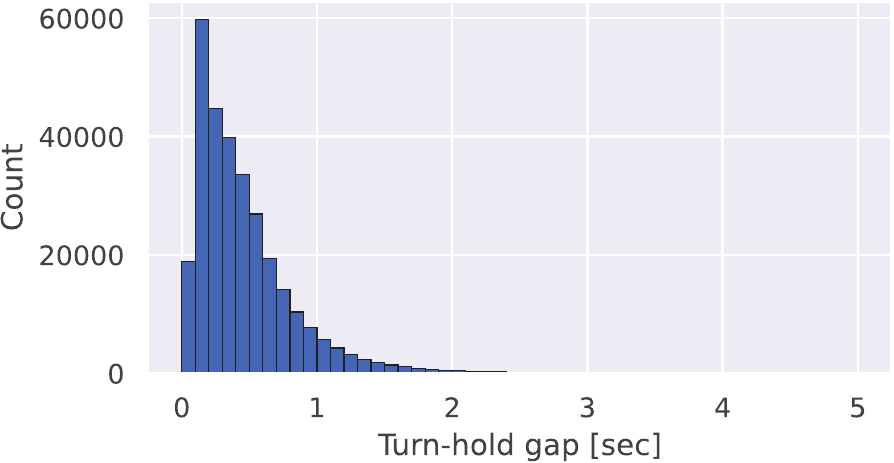}
        \caption{English (Switchboard)}
    \end{subfigure}
    \hfill
    \begin{subfigure}{0.32\textwidth}
    \centering
        \includegraphics[width=\linewidth]{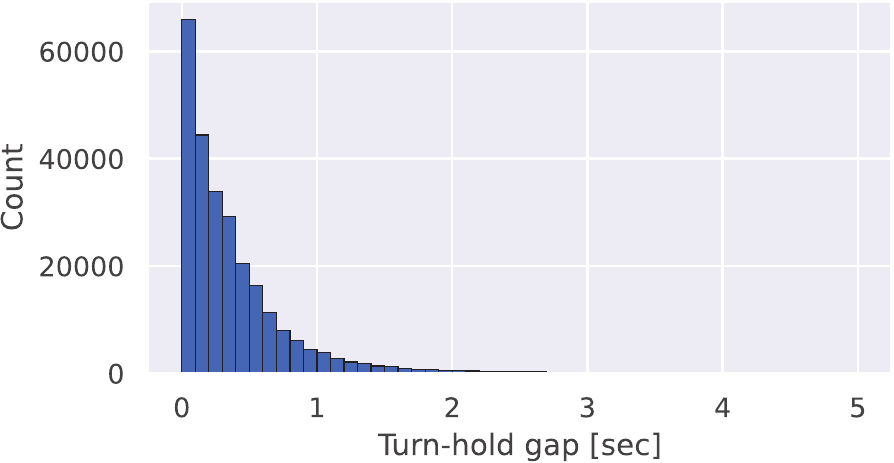}
        \caption{Mandarin (HKUST)}
    \end{subfigure}
    \hfill
    \begin{subfigure}{0.32\textwidth}
    \centering
        \includegraphics[width=\linewidth]{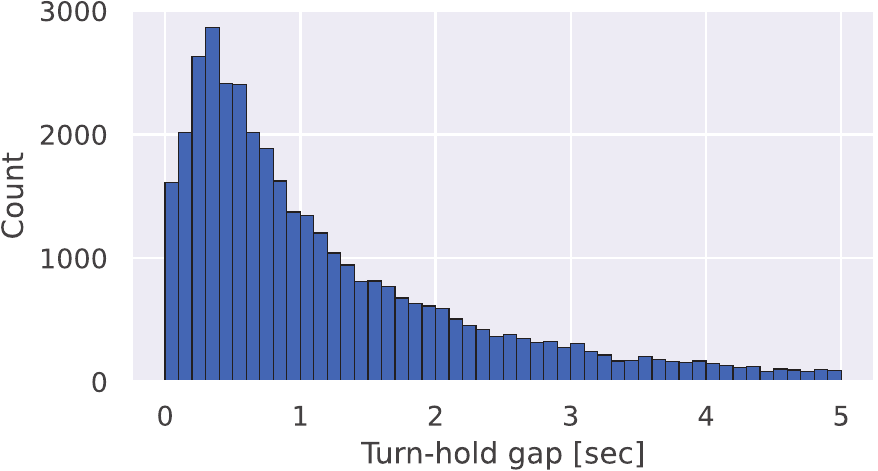}
        \caption{Japanese (Travel agency)}
    \end{subfigure}
    \caption{Histogram of \textbf{turn-hold} gap in three languages}
    \label{fig:hist:hold}
\end{figure*}

\section{Datasets} \label{sec:dataset}

In this study, we use three dyadic conversational datasets: English, Mandarin, and Japanese, to investigate a model for multilingual turn-taking.

\subsection{Switchboard (English)}

The Switchboard dataset is a collection of telephone conversations recorded in English, covering everyday topics~\cite{swb}.
It consists of a total of 2,438 dialogues, equivalent to approximately 259.1 hours of data.
This dataset was divided into training, validation, and test sets in a ratio of 8:1:1 at the session level, using random selection.
The training set comprises 1950 dialogues (218.7 hours).
The validation and test sets contain 244 dialogues (20.2 hours), respectively.
Since not all datasets were of equal size, and in order to make the comparison between languages fair, we selected a subset of the data to be used. 
Therefore, the training and validation sets were further randomly sub-sampled to approximately 92.5 and 11.5 hours, in order to align them to the size of the smallest dataset, namely the Japanese dataset.

\subsection{HKUST Mandarin Telephone Speech}

The HKUST Mandarin telephone speech corpus is a collection of Mandarin Chinese spoken dialogues in telephone conversations~\cite{hkust}, similar to Switchboard.
The dataset contains a total of 867 dialogues, approximately 148.6 hours in duration.
These have been divided into the training of 758 dialogues (approximately 130.1 hours), the validation of 88 dialogues (14.5 hours), and the test of 24 dialogues (3.9 hours), respectively.
Similar to Switchboard, in order to match the size of the smallest dataset, which is the Japanese dataset, the training and validation sets have been sub-sampled to 92.5 hours and 11.5 hours, respectively.

\subsection{Travel Agency Task Dialogues (Japanese)}

Travel Agency Task Dialogues is a project that collects simulated dialogue data for travel consultations in Japanese~\cite{inaba2023travel}.
These conversations were recorded using an online conference system.
The dialogues simulate online conversations between a travel agency staff and a customer, with the role of the staff being played by someone with actual experience working in a travel agency.
Note that while the participants were assigned roles, they were not given a script.
A total of 329 dialogues (115.5 hours) have been recorded.
These dialogues were randomly divided into session units, with 263 dialogues (92.5 hours) in the training set, and 33 dialogues (11.5 hours) each in the validation set and test set.

\subsection{Differences Across Languages}

As mentioned above, earlier studies suggest that there are slight differences in turn-taking tendencies between the three languages.
Therefore, we first investigated the differences in gap and pause length between the datasets.
Figure~\ref{fig:hist:gap} shows the histogram of gap duration during turn transitions, while Figure~\ref{fig:hist:hold} illustrates the distribution of pause length during turn holding.
It can be observed that Mandarin and Japanese tend to have slightly shorter gaps during turn transitions compared to English.
On the other hand, during turn holding, English and Mandarin show a tendency towards shorter pause lengths, compared to Japanese.
Furthermore, it's worth observing that the distribution of Japanese is more evenly dispersed.
The Japanese dataset exhibits a formal dialogue setting with explicitly assigned roles, distinguishing it from the other two datasets.
Consequently, there seems to be a trend in the Japanese dataset where the participants hold the turn for longer periods of time without turn-shifts, resulting in longer gaps between utterances.

Given these differences, it is important to note that the VAP model does not only have to take into account how cues and signals may differ between languages, but also the overall distributions, which might bias the predictions. 

\section{Experiments} \label{sec:exp}

To answer the research questions from \textbf{RQ1} to \textbf{RQ5} mentioned in Section~\ref{sec:intro}, we conducted a series of experiments, as described below.

\subsection{Cross-lingual Performance} \label{sec:exp:cross}

In order to answer \textbf{RQ1} and \textbf{RQ2}, we compared the performances between a multilingual model and monolingual models trained specifically for each language.

\subsubsection{Condition}

The multilingual model was trained on all the data from the three languages mentioned above.
Additionally, a monolingual model was trained separately for each language.
Thus, the training data quantity of the multilingual model was three times that of each monolingual model.

The structure and parameters of the VAP model were the same for the multilingual and monolingual settings.
The training parameters were as follows: The number of training epochs was 20, batch size was 8, learning rate was 3.63E-4, and weight decay was set to 0.001. We used the AdamW optimizer.
We evaluated the test set using the model with the smallest loss on the validation set.

\subsubsection{Test Loss Performance}

As a basic evaluation metric, we assessed the average loss on the test set.
The loss of interest for evaluation is $L_{vap}$, which was defined in Section~\ref{sec:vap}.
Table~\ref{table:cross:testloss} shows the results. As can be seen, whereas the monolingual models work well when tested on the same language, they perform considerably worse when applied to another language.
From this result, it is clear that the nature of voice activity projection differs across the three language datasets used in this study, and in order to make accurate predictions, it is necessary to train specific models for each language.
However, the results of the multilingual model reveal that it can project voice activity with the same level of performance for all languages as the language-matched models.

\begin{table}[t]
    \centering
    \tabcolsep=3mm
    \begin{tabular}{lccc}
        \hline
        & \multicolumn{3}{c}{Test data} \\
        \cline{2-4}
        \multicolumn{1}{c}{Training data} & ENG & MAN & JPN \\
        \hline
        English & \textbf{2.387}	& 3.401	& 2.956 \\
        Mandarin & 2.839 & \textbf{2.817} & 3.098 \\
        Japanese & 3.306 & 4.004 & 2.329 \\
        Multi (proposed) & 2.396 & 2.832 & \textbf{2.265} \\
        \hline
    \end{tabular}
    \caption{Test loss on cross-lingual performance}
    \label{table:cross:testloss}
\end{table}

\subsubsection{Turn Shift/hold Prediction} \label{sec:exp1:turn}

While the test loss reveals the general performance of the models, the numbers are hard to interpret. Thus, we also evaluated the applicability of a multilingual VAP model in the typical problem of predicting a \textit{shift} vs. a \textit{hold} in periods of mutual silence. This is an application that is similar to end-of-turn prediction in spoken dialogue systems.
This evaluation is the same as the one used in previous research~\cite{erik2022vap}.
The task is to predict whether the preceding speaker and the following speaker were different (shift) or the same (hold) when a mutual silence longer than 0.25 seconds is observed.
Note that the preceding and following utterances must be longer than one second.
The value of $p_{now}$ after 0.05 seconds into the start of the mutual silence is used to predict who the next speaker is.

The distribution of the shift/hold classes is summarized in Table~\ref{table:cross:dist-shift-hold}.
Since the nature of the dialogue varies depending on the language, the ratio of turn shifts to holds is also different.
For example, in the Japanese data, there are fewer pauses (holds) and more concise utterances.
The Mandarin data has the next largest imbalance, followed by English.
In particular, in English, holds are approximately 10 times more frequent than shifts.

Although the evaluation metric used in previous research was the weighted F1 score, in this study, we used balanced accuracy to reduce the bias of class imbalance between languages.
Furthermore, since the value of balanced accuracy would be 0.5 for random or majority-class prediction, it also has the advantage of being easily interpretable.

Table~\ref{table:cross:result-shift-hold} shows the prediction results.
The results are in line with those obtained using the test loss. 
Furthermore, when comparing the accuracy across languages, it is evident that Mandarin has the most predictable turn-taking patterns.
From these results, we conclude that the monolingual models cannot be directly applied to other languages (\textbf{RQ1}), but that the multilingual model can be utilized as a generic turn-taking model for all three languages (\textbf{RQ2}).

\begin{figure*}[t]
    \centering
    (a) English (Switchboard)
    \includegraphics[width=\linewidth]{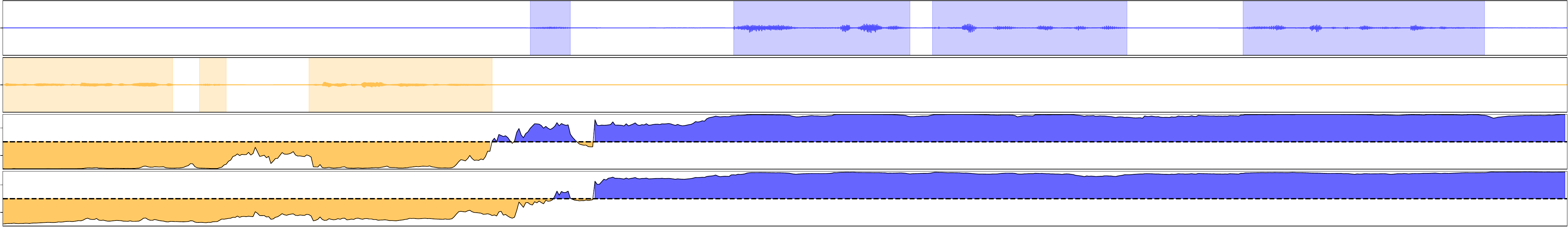}
    \vspace{-1mm} \\
    (b) Mandarin (HKUST)
    \includegraphics[width=\linewidth]{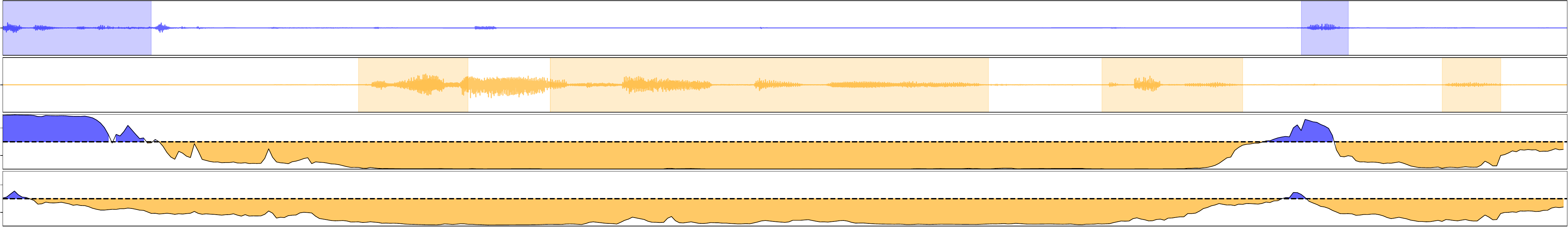}
    \vspace{-1mm} \\
    (c) Japanese (Travel agency)
    \includegraphics[width=\linewidth]{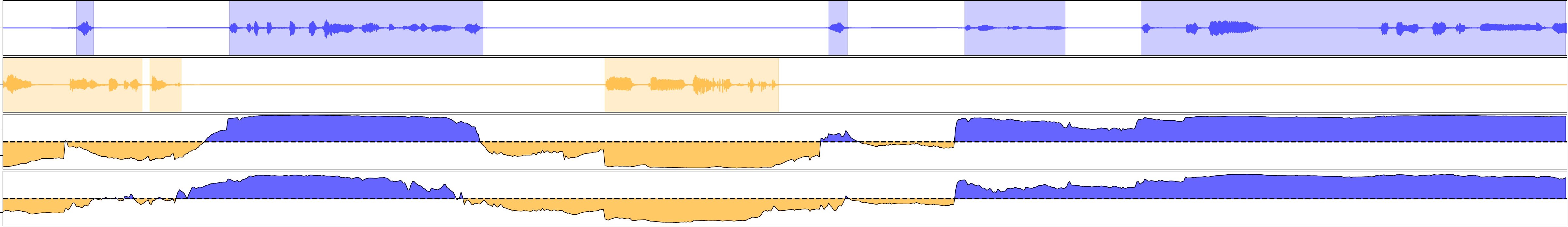}
    \caption{Output example of multilingual VAP in three languages (Top: English, Middle: Mandarin, Bottom: Japanese) - Each graph consists of, from top to bottom, input waveforms of both participants, near future voiced probability ($p_{now}$), and future voiced probability ($p_{future}$) among participants.}
    \label{fig:exp1:example1}
\end{figure*}

\begin{table}[t]
    \centering
    \tabcolsep=3mm
    \begin{tabular}{lrrr}
        \hline
        \multicolumn{1}{c}{Dataset} & \#Shift & \#Hold & \%Shift \\
        \hline
        English & 1253	& 11432 & 9.9\\
        Mandarin & 718 & 1807 & 28.4 \\
        Japanese & 1029 & 1371 & 42.9 \\
        \hline
    \end{tabular}
    \caption{Distribution of samples for turn shift/hold prediction}
    \label{table:cross:dist-shift-hold}
\end{table}

\begin{table}[t]
    \centering
    \tabcolsep=3mm
    \begin{tabular}{lccc}
        \hline
        \multicolumn{1}{c}{\multirow{2}{*}{Training data}} & \multicolumn{3}{c}{Test data} \\
        \cline{2-4}
         & ENG & MAN & JPN \\
        \hline
        English & \textbf{79.59}	& 68.64	& 59.43 \\
        Mandarin & 65.31 & 84.49 & 59.72 \\
        Japanese & 64.46 & 67.89 & 74.20 \\
        Multi (proposed) & 77.16 & \textbf{84.60} & \textbf{76.54} \\
        \hline
    \end{tabular}
    \caption{Cross-lingual turn shift/hold prediction performance (balanced accuracy [\%])}
    \label{table:cross:result-shift-hold}
\end{table}

Figure~\ref{fig:exp1:example1} illustrates output examples of the multilingual model in each language dataset.
In these figures, both $p_{now}$ and $p_{future}$ are illustrated together with input waveforms colored with reference VAD segments.
In the English example (a), the turn shifts from the orange participant to the blue participant.
During this turn shift, $p_{now}$ exhibits a notably high predictive value before blue's speech. Additionally, during the pauses within each speaker's turn (holds), $p_{now}$ and $p_{future}$ correctly predict a continuation of each speaker's turn.

This same pattern is observed in the Mandarin example (b).
In this case, we can see how $p_{future}$ projects a turn-completion already before blue has stopped speaking. 
Furthermore, in the latter part of the Mandarin example, there is a short backchannel from blue, which is predicted by $p_{now}$.
As this value is stronger than $p_{future}$, this illustrates how $p_{now}$ and $p_{future}$ can be used together to predict backchannels, but also for predicting that blue will not produce a longer utterance, after the onset of the backchannel.

The Japanese example (c) also illustrates that both turn shifts and holds can be predicted effectively.
Note that there is a prolonged pause at the beginning of blue's turn-taking around the middle, during which $p_{now}$ and $p_{future}$ demonstrate an uncertainty as to who will be the next speaker.
These types of situations are commonly observed in natural conversations, and are referred to as \textit{self-selection} in the literature~\cite{sacks1978simplest}. This kind of information could be utilized by spoken dialogue systems, allowing it to either take the turn or leave it to the user.

\subsection{Language Identification} \label{sec:exp:lid}

Next, to answer \textbf{RQ3}, we investigate the language identification ability of the multilingual model.
Based on the observation that the multilingual model performed well in all three languages in the previous section, and that the monolingual models did not perform well for other languages, we hypothesized that the multilingual model is able to identify the language of the input speech and operates accordingly.
To investigate this, we added another linear layer for language identification to the final layer of the VAP model, along with those for VAP and VAD.
Since we are dealing with three languages, it becomes a three-class classification problem.
Then, we added the cross-entropy-based language identification loss ($L_{lid}$) to the training loss as
\begin{equation}
    L = L_{vap} + L_{vad} + L_{lid} ~ .
\end{equation}
We then trained this model from scratch in this experiment.

As a result, when the language identification accuracy was measured on the test set, it reached a weighted F1-score of 99.99\%.
In other words, the multilingual model is able to almost perfectly identify the language of the input speech. 

We also wanted to see whether this added language identification loss would act as a multi-task loss, potentially improving the performance of the multi-lingual model.
Table~\ref{table:lid} reports the performance on the VAP test loss ($L_{vap}$) and the balanced accuracy of shift/hold prediction for turn-taking, for this new model compared to the previous model. As the results are very similar between models, we draw the conclusion that the model does not need help to learn to identify the language, but that it does so anyway implicitly.

\begin{table}[t]
    \centering
    \tabcolsep=1.5mm
    \begin{tabular}{lccccc}
        \hline
        \multicolumn{1}{c}{\multirow{2}{*}{Test data}} & \multicolumn{2}{c}{Test loss ($\downarrow$)} & & \multicolumn{2}{c}{Shift/Hold ($\uparrow$)} \\
        \cline{2-3}
        \cline{5-6}
        & w/o LID & w/ LID  & & w/o LID & w/ LID \\
        \hline
        English  & 2.396 & 2.401 & & 79.59 & 78.44 \\
        Mandarin & 2.832 & 2.819 & & 84.49 & 84.72 \\
        Japanese & 2.265 & 2.341 & & 74.20 & 75.82 \\
        \hline
    \end{tabular}
    \caption{Performance with or without language identification (LID) multitask (``Shift/Hold'' represents the balanced accuracy [\%] on the turn shift/hold prediction task.)}
    \label{table:lid}
\end{table}

\subsection{Pitch Sensitivity} \label{sec:exp:pitch}

\begin{figure}[t]
  \begin{minipage}{0.49\linewidth}
    \centering
    Original (before)
    \includegraphics[width=\linewidth]{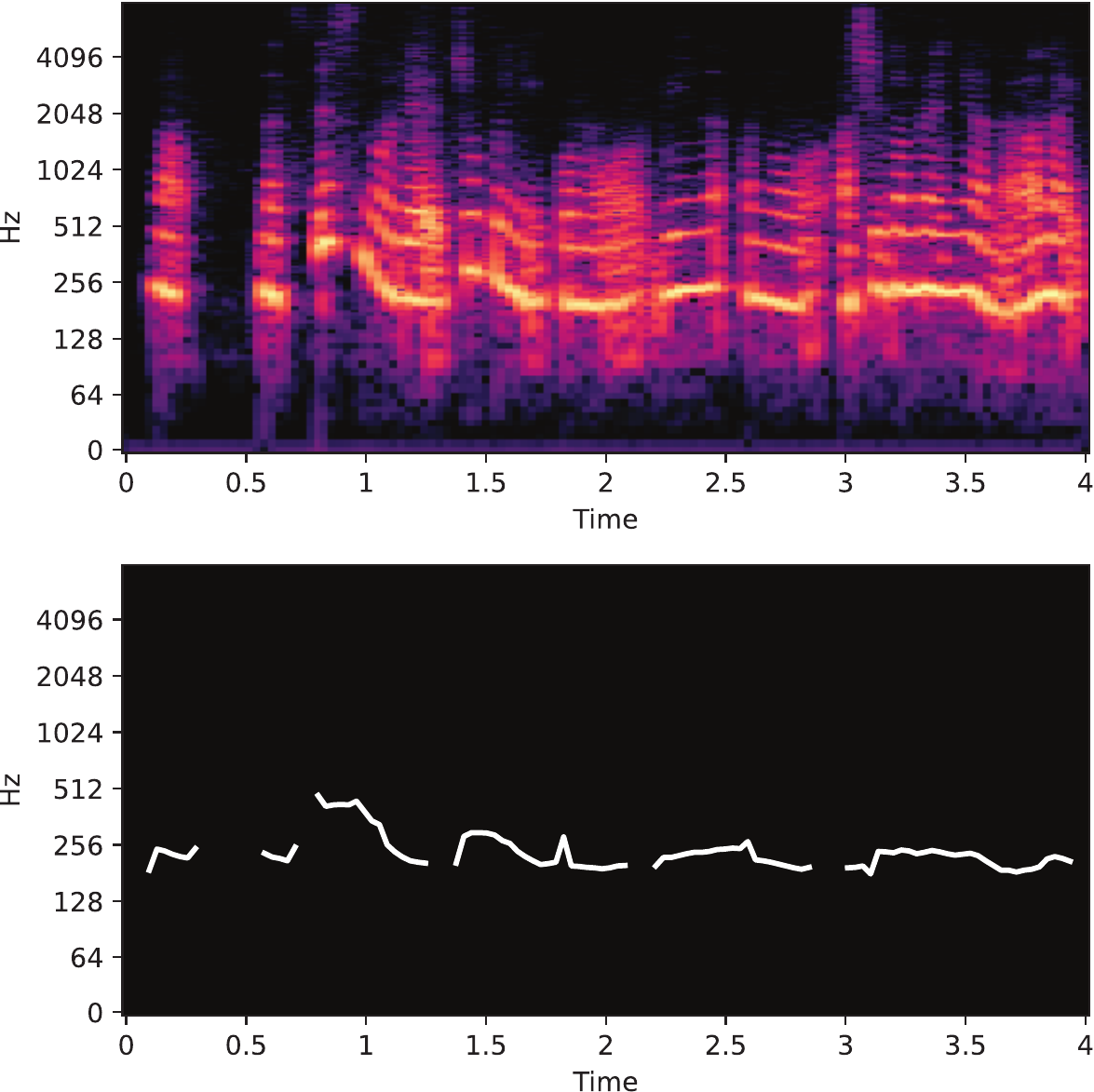}
  \end{minipage}
  \begin{minipage}{0.49\linewidth}
    \centering
    Pitch flattened (after)
    \includegraphics[width=\linewidth]{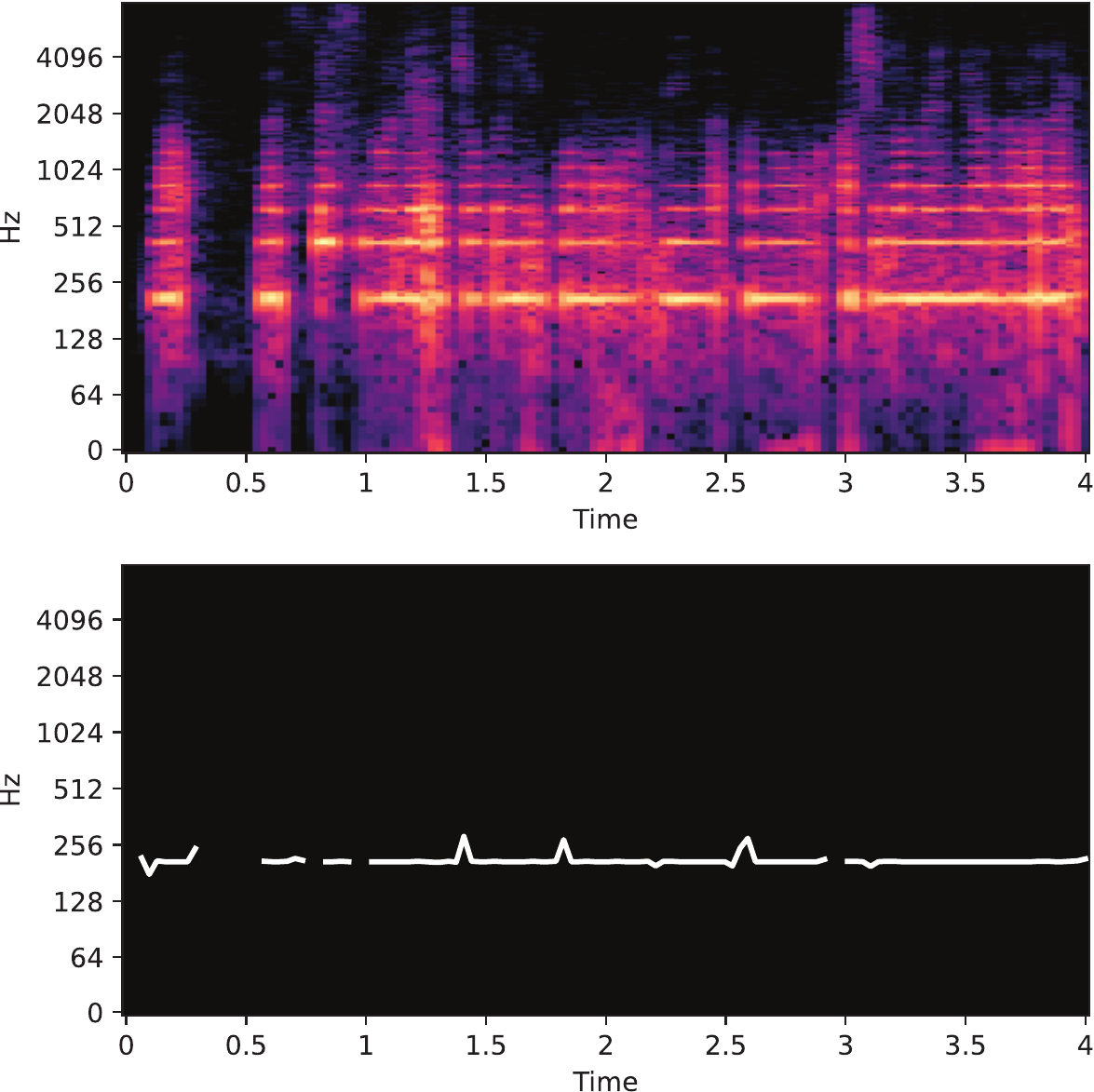}
  \end{minipage}
  \caption{Input example of pitch flattening test (Top: Spectrogram, Bottom: Automatically estimated F0)}
  \label{fig:exp:pitch_flatten}
\end{figure}

As mentioned in Section~\ref{sec:intro}, prosodic information is an important factor in predicting turn-taking.
To assess the model's reliance on prosodic information indirectly and to answer \textbf{RQ4}, we flattened the pitch of the input speech during the test phase (Figure~\ref{fig:exp:pitch_flatten}) and measured the resulting performance degradation.
A previous study conducted a similar test and found that pitch flattening had a minor overall impact on the model's performance, while being important in specific contexts of syntactic ambiguity~\cite{erik2022sigdial}.
In this study, we used Praat\footnote{\url{https://www.fon.hum.uva.nl/praat/}} to flatten the pitch, following the methodology of the aforementioned study.
We then examined the performance changes before and after pitch flattening for the turn shift/hold prediction.

\begin{table}[t]
    \centering
    \tabcolsep=2mm
    \begin{tabular}{lcc}
        \hline
        Test data & Mono & Multi \\
        \hline
        English  & 79.68 ($+$0.09) & 76.28 ($+$0.12) \\
        Mandarin & 82.47 ($-$2.02) & 82.30 ($-$2.30) \\
        Japanese & 72.83 ($-$1.37) & 74.73 ($-$1.81) \\
        \hline
    \end{tabular}
    \caption{Turn shift/hold prediction performance (balanced accuracy [\%]) with pitch flattening (difference against the case without pitch flattening)}
    \label{table:exp:pitch}
\end{table}

Table~\ref{table:exp:pitch} presents the results for both monolingual and multilingual models.
Both types of models exhibited similar trends, indicating that they rely on pitch information to a similar extent~(\textbf{RQ4}).
Furthermore, when analyzing the differences between languages, a decrease in accuracy of approximately two percentage units was observed in Japanese and Mandarin.
In contrast, the change in accuracy for English was not significant.
This discrepancy suggests that turn-taking in Japanese and Mandarin is more dependent on pitch cues, compared to English.

For Mandarin, together with the high accuracy in the shift/hold prediction experiment, our results suggest that pitch plays an important role for indicating turn-final position, which aligns with previous findings.
While falling intonation is a universal turn-yielding cue in various languages~\cite{mccarthy1991discourse}, Mandarin, being a tonal language with lexical tones for individual words, may be able to provide richer pitch information for the model.
For example, \citet{levow2005turn} investigated the interaction between tone and intonation and revealed that the pitch of turn-final words is nevertheless relatively lower than those in other positions in the utterance.
They further show that intonation patterns can be used to train a classifier for determining turn-final syllables.

\subsection{Effect of Audio Encoder} \label{sec:exp:mms}

Finally, we also investigated the audio encoder used in the VAP model to answer~\textbf{RQ5}.
The encoder we used is CPC, which was also used in previous research on the VAP model.
One potential limiting factor in a multilingual setting is that the model is trained on the English Librispeech dataset.
Additionally, during the training of the VAP model, the parameters of this CPC are frozen. On the other hand, according to previous research, the English CPC model has been shown to be effective for phoneme recognition in other languages~\cite{riviere2020unsupervised}.
It should be noted that it is not straightforward to adopt any audio encoder for the VAP model, since the encoder needs to operate in a causal manner. This rules out models such as HuBERT~\cite{hsu2021hubert}, which is bidirectional. 

Still, we wanted to compare it with an audio encoder that is multilingual and not pre-trained on a specific language. Recently, Meta has released a multilingual wav2vec 2.0 model called Massively Multilingual Speech (MMS), which is pre-trained on data from 1406 languages~\cite{pratap2023mms}. Although the transformer layers of wav2vec 2.0 are bidirectional, the initial multi-stage CNN operates on a much smaller time window. Thus, we could adopt this initial part of the model for the VAP model, and compare the performance with CPC. 
Just like with CPC, this multi-stage CNN was frozen during the training of the VAP model.

The comparative results are reported in Table~\ref{table:encoder}.
Overall, MMS results show slightly lower performance compared to CPC.
Thus, CPC seems to be more compatible with the task of the VAP model~(\textbf{RQ5}).
We also tried to train the entire model without freezing the audio encoder. However, both CPC and MMS showed a slight decrease in accuracy in doing so.
Given the current size of the training dataset, there is a possibility that the model is overfitting.

\begin{table}[t]
    \centering
    \tabcolsep=2mm
    \begin{tabular}{lccccc}
        \hline
        \multicolumn{1}{c}{\multirow{2}{*}{Test data}} & \multicolumn{2}{c}{Test loss ($\downarrow$)} & & \multicolumn{2}{c}{Shift/Hold ($\uparrow$)} \\
        \cline{2-3}
        \cline{5-6}
        & CPC & MMS  & & CPC & MMS \\
        \hline
        English  & 2.396 & 2.421 & & 79.59 & 77.67 \\
        Mandarin & 2.832 & 2.841 & & 84.49 & 82.09 \\
        Japanese & 2.265 & 2.394 & & 74.20 & 72.10 \\
        \hline
    \end{tabular}
    \caption{Performance on comparison of audio encoders (``Shift/Hold'' reports balanced accuracy [\%] on the turn shift/hold prediction task.)}
    \label{table:encoder}
\end{table}

\section{Conclusion} \label{sec:conclusion}

In this paper, we have investigated the application of voice activity projection (VAP), a predictive turn-taking model for spoken dialogue, on multilingual data, encompassing English, Mandarin and Japanese. The results show that a monolingual VAP model does not work well when applied to other languages~(\textbf{RQ1}). However, a multilingual VAP model (trained on all languages) shows comparable performance to monolingual models across all three language datasets~(\textbf{RQ2}).
Next, by incorporating an additional language identification task, we showed that the multilingual model can accurately identify the language of input audio (\textbf{RQ3}).
We also investigated the model's sensitivity to pitch, by flattening the pitch of the input audio, Whereas the performance on English did not change by this perturbation, Japanese and Mandarin seem to be somewhat more dependent on pitch cues~(\textbf{RQ4}).
Finally, with respect to the audio encoder, we have found that the current pre-trained CPC model is better than the other alternative we have tried~(\textbf{RQ5}).

\section*{Acknowledgements}

This work was supported by JST ACT-X JPMJAX2103, JST Moonshot R\&D JPMJPS2011, JSPS KAKENHI JP23K16901, Riksbankens Jubileumsfond (RJ) P20-0484, and Swedish Research Council (VR) 2020-03812.

\section*{References} \label{sec:reference}
\bibliographystyle{lrec-coling2024-natbib}
\bibliography{main}

\begin{thebibliography}{41}
\expandafter\ifx\csname natexlab\endcsname\relax\def\natexlab#1{#1}\fi

\bibitem[{Clancy et~al.(1996)Clancy, Thompson, Suzuki, and Tao}]{clancy1996conversational}
Patricia~M Clancy, Sandra~A Thompson, Ryoko Suzuki, and Hongyin Tao. 1996.
\newblock The conversational use of reactive tokens in {English}, {Japanese}, and {Mandarin}.
\newblock \emph{Journal of pragmatics}, 26(3):355--387.

\bibitem[{Dingemanse and Liesenfeld(2022)}]{dingemanse2022text}
Mark Dingemanse and Andreas Liesenfeld. 2022.
\newblock From text to talk: {Harnessing} conversational corpora for humane and diversity-aware language technology.
\newblock In \emph{Annual Meeting of the Association for Computational Linguistics (ACL)}, pages 5614--5633.

\bibitem[{Duncan(1972)}]{duncan1972some}
Starkey Duncan. 1972.
\newblock Some signals and rules for taking speaking turns in conversations.
\newblock \emph{Journal of personality and social psychology}, 23(2):283--292.

\bibitem[{Ekstedt and Skantze(2020)}]{ekstedt2020turngpt}
Erik Ekstedt and Gabriel Skantze. 2020.
\newblock {TurnGPT:} {A} {Transformer-based} language model for predicting turn-taking in spoken dialog.
\newblock In \emph{Empirical Methods in Natural Language Processing (EMNLP)}, pages 2981--2990.

\bibitem[{Ekstedt and Skantze(2022{\natexlab{a}})}]{erik2022sigdial}
Erik Ekstedt and Gabriel Skantze. 2022{\natexlab{a}}.
\newblock How much does prosody help turn-taking? {Investigations} using voice activity projection models.
\newblock In \emph{Annual Meeting of the Special Interest Group on Discourse and Dialogue (SIGdial)}, pages 541--551.

\bibitem[{Ekstedt and Skantze(2022{\natexlab{b}})}]{erik2022vap}
Erik Ekstedt and Gabriel Skantze. 2022{\natexlab{b}}.
\newblock {Voice Activity Projection}: {Self-supervised} learning of turn-taking events.
\newblock In \emph{INTERSPEECH}, pages 5190--5194.

\bibitem[{Garrod and Pickering(2015)}]{garrod2015use}
Simon Garrod and Martin~J Pickering. 2015.
\newblock The use of content and timing to predict turn transitions.
\newblock \emph{Frontiers in psychology}, 6(751):1--12.

\bibitem[{Godfrey et~al.(1992)Godfrey, Holliman, and McDaniel}]{swb}
John~J Godfrey, Edward~C Holliman, and Jane McDaniel. 1992.
\newblock {SWITCHBOARD:} {Telephone} speech corpus for research and development.
\newblock In \emph{International Conference on Acoustics, Speech, and Signal Processing (ICASSP)}, pages 517--520.

\bibitem[{Gravano and Hirschberg(2011)}]{gravano2011turn}
Agust{\'\i}n Gravano and Julia Hirschberg. 2011.
\newblock Turn-taking cues in task-oriented dialogue.
\newblock \emph{Computer Speech \& Language}, 25(3):601--634.

\bibitem[{Heldner and Edlund(2010)}]{heldner2010pauses}
Mattias Heldner and Jens Edlund. 2010.
\newblock Pauses, gaps and overlaps in conversations.
\newblock \emph{Journal of Phonetics}, 38(4):555--568.

\bibitem[{Hsu et~al.(2021)Hsu, Bolte, Tsai, Lakhotia, Salakhutdinov, and Mohamed}]{hsu2021hubert}
Wei-Ning Hsu, Benjamin Bolte, Yao-Hung~Hubert Tsai, Kushal Lakhotia, Ruslan Salakhutdinov, and Abdelrahman Mohamed. 2021.
\newblock {HuBERT:} {Self-supervised} speech representation learning by masked prediction of hidden units.
\newblock \emph{IEEE/ACM Transactions on Audio, Speech, and Language Processing}, 29:3451--3460.

\bibitem[{Inaba et~al.(2022)Inaba, Chiba, Higashinaka, Komatani, Miyao, and Nagai}]{inaba2023travel}
Michimasa Inaba, Yuya Chiba, Ryuichiro Higashinaka, Kazunori Komatani, Yusuke Miyao, and Takayuki Nagai. 2022.
\newblock Collection and analysis of travel agency task dialogues with age-diverse speakers.
\newblock In \emph{Language Resources and Evaluation Conference (LREC)}, pages 5759--5767.

\bibitem[{Inoue et~al.(2024)Inoue, Jiang, Ekstedt, Kawahara, and Skantze}]{inoue2024iwsds}
Koji Inoue, Bing'er Jiang, Erik Ekstedt, Tatsuya Kawahara, and Gabriel Skantze. 2024.
\newblock Real-time and continuous turn-taking prediction using voice activity projection.
\newblock In \emph{International Workshop on Spoken Dialogue Systems Technology (IWSDS)}.

\bibitem[{Ishimoto et~al.(2017)Ishimoto, Teraoka, and Enomoto}]{ishimoto2017interspeech}
Yuichi Ishimoto, Takehiro Teraoka, and Mika Enomoto. 2017.
\newblock End-of-utterance prediction by prosodic features and phrase-dependency structure in spontaneous {Japanese} speech.
\newblock In \emph{INTERSPEECH}, pages 1681--1685.

\bibitem[{Jian and Wu(2011)}]{jian2011mandarin}
Hua-Li Jian and Joyce Wu. 2011.
\newblock Mandarin conversation: {Turn-taking} cues in exchange structure.
\newblock In \emph{International Congress of Phonetic Sciences (ICPhS)}, pages 970--973.

\bibitem[{Kendrick et~al.(2023)Kendrick, Holler, and Levinson}]{kendrick2023turn}
Kobin~H Kendrick, Judith Holler, and Stephen~C Levinson. 2023.
\newblock Turn-taking in human face-to-face interaction is multimodal: {Gaze} direction and manual gestures aid the coordination of turn transitions.
\newblock \emph{Philosophical Transactions of the Royal Society B}, 378(1875):20210473.

\bibitem[{Koiso et~al.(1998)Koiso, Horiuchi, Tutiya, Ichikawa, and Den}]{koiso1998analysis}
Hanae Koiso, Yasuo Horiuchi, Syun Tutiya, Akira Ichikawa, and Yasuharu Den. 1998.
\newblock An analysis of turn-taking and backchannels based on prosodic and syntactic features in {Japanese} map task dialogs.
\newblock \emph{Language and speech}, 41(3-4):295--321.

\bibitem[{Kurata et~al.(2023)Kurata, Saeki, Fujie, and Matsuyama}]{kurata23_interspeech}
Fuma Kurata, Mao Saeki, Shinya Fujie, and Yoichi Matsuyama. 2023.
\newblock Multimodal turn-taking model using visual cues for end-of-utterance prediction in spoken dialogue systems.
\newblock In \emph{INTERSPEECH}, pages 2658--2662.

\bibitem[{Lala et~al.(2018)Lala, Inoue, and Kawahara}]{lala2018icmi}
Divesh Lala, Koji Inoue, and Tatsuya Kawahara. 2018.
\newblock Evaluation of real-time deep learning turn-taking models for multiple dialogue scenarios.
\newblock In \emph{International Conference on Multimodal Interaction (ICMI)}, pages 78--86.

\bibitem[{Lala et~al.(2019)Lala, Inoue, and Kawahara}]{lala2019icmi}
Divesh Lala, Koji Inoue, and Tatsuya Kawahara. 2019.
\newblock Smooth turn-taking by a robot using an online continuous model to generate turn-taking cues.
\newblock In \emph{International Conference on Multimodal Interaction (ICMI)}, pages 226--234.

\bibitem[{Lala et~al.(2017)Lala, Milhorat, Inoue, Ishida, Takanashi, and Kawahara}]{lala2017sigdial}
Divesh Lala, Pierrick Milhorat, Koji Inoue, Masanari Ishida, Katsuya Takanashi, and Tatsuya Kawahara. 2017.
\newblock Attentive listening system with backchanneling, response generation and flexible turn-taking.
\newblock In \emph{Annual Meeting of the Special Interest Group on Discourse and Dialogue (SIGdial)}, pages 127--136.

\bibitem[{Levinson and Torreira(2015)}]{Levinson2015TurnTaking}
Stephen~C. Levinson and Francisco Torreira. 2015.
\newblock Timing in turn-taking and its implications for processing models of language.
\newblock \emph{Frontiers in Psychology}, 6(731):1--17.

\bibitem[{Levow(2005)}]{levow2005turn}
Gina-Anne Levow. 2005.
\newblock Turn-taking in {Mandarin} dialogue: {Interactions} of tone and intonation.
\newblock In \emph{SIGHAN Workshop on Chinese Language Processing (SIGHAN)}.

\bibitem[{Liermann et~al.(2023)Liermann, Park, Choi, and Lee}]{liermann2023emnlp}
Wencke Liermann, Yo-Han Park, Yong-Seok Choi, and Kong Lee. 2023.
\newblock Dialogue act-aided backchannel prediction using multi-task learning.
\newblock In \emph{Empirical Methods in Natural Language Processing (EMNLP)}, pages 15073--15079.

\bibitem[{Liu et~al.(2006)Liu, Fung, Yang, Cieri, Huang, and Graff}]{hkust}
Yi~Liu, Pascale Fung, Yongsheng Yang, Christopher Cieri, Shudong Huang, and David Graff. 2006.
\newblock {HKUST/MTS:} {A} very large scale mandarin telephone speech corpus.
\newblock In \emph{International Symposium Chinese Spoken Language Processing (ISCSLP)}, pages 724--735.

\bibitem[{Local et~al.(1986)Local, Kelly, and Wells}]{local1986towards}
John~K Local, John Kelly, and William~HG Wells. 1986.
\newblock Towards a phonology of conversation: turn-taking in tyneside english1.
\newblock \emph{Journal of Linguistics}, 22(2):411--437.

\bibitem[{Masumura et~al.(2017)Masumura, Asami, Masataki, Ishii, and Higashinaka}]{masumura2017}
Ryo Masumura, Taichi Asami, Hirokazu Masataki, Ryo Ishii, and Ryuichiro Higashinaka. 2017.
\newblock Online end-of-turn detection from speech based on stacked time-asynchronous sequential networks.
\newblock In \emph{INTERSPEECH}, pages 1661--1665.

\bibitem[{McCarthy(1991)}]{mccarthy1991discourse}
Michael McCarthy. 1991.
\newblock \emph{Discourse analysis for language teachers}.
\newblock Cambridge university press.

\bibitem[{Muromachi and Kano(2023)}]{muromachi2023interspeech}
Toshiki Muromachi and Yoshinobu Kano. 2023.
\newblock {Estimation of Listening Response Timing by Generative Model and Parameter Control of Response Substantialness Using Dynamic-Prompt-Tune}.
\newblock In \emph{INTERSPEECH}, pages 2638--2642.

\bibitem[{Nguyen et~al.(2023)Nguyen, Kharitonov, Copet, Adi, Hsu, Elkahky, Tomasello, Algayres, Sagot, Mohamed et~al.}]{nguyen2023generative}
Tu~Anh Nguyen, Eugene Kharitonov, Jade Copet, Yossi Adi, Wei-Ning Hsu, Ali Elkahky, Paden Tomasello, Robin Algayres, Benoit Sagot, Abdelrahman Mohamed, et~al. 2023.
\newblock Generative spoken dialogue language modeling.
\newblock \emph{Transactions of the Association for Computational Linguistics}, 11:250--266.

\bibitem[{Onishi et~al.(2023)Onishi, Tanaka, and Nakamura}]{onishi2023hai}
Kazuyo Onishi, Hiroki Tanaka, and Satoshi Nakamura. 2023.
\newblock Multimodal voice activity prediction: {Turn-taking} events detection in expert-novice conversation.
\newblock In \emph{International Conference on Human-Agent Interaction (HAI)}, pages 13--21.

\bibitem[{Pratap et~al.(2023)Pratap, Tjandra, Shi, Tomasello, Babu, Kundu, Elkahky, Ni, Vyas, Fazel-Zarandi, Baevski, Adi, Zhang, Hsu, Conneau, and Auli}]{pratap2023mms}
Vineel Pratap, Andros Tjandra, Bowen Shi, Paden Tomasello, Arun Babu, Sayani Kundu, Ali Elkahky, Zhaoheng Ni, Apoorv Vyas, Maryam Fazel-Zarandi, Alexei Baevski, Yossi Adi, Xiaohui Zhang, Wei-Ning Hsu, Alexis Conneau, and Michael Auli. 2023.
\newblock Scaling speech technology to 1,000+ languages.
\newblock \emph{arXiv preprint}.
\newblock ArXiv:2305.13516.

\bibitem[{Raux and Eskenazi(2012)}]{raux2012}
Antoine Raux and Maxine Eskenazi. 2012.
\newblock Optimizing the turn-taking behavior of task-oriented spoken dialog systems.
\newblock \emph{ACM Transactions on Speech and Language Processing}, 9(1):1--23.

\bibitem[{Riviere et~al.(2020)Riviere, Joulin, Mazar{\'e}, and Dupoux}]{riviere2020unsupervised}
Morgane Riviere, Armand Joulin, Pierre-Emmanuel Mazar{\'e}, and Emmanuel Dupoux. 2020.
\newblock Unsupervised pretraining transfers well across languages.
\newblock In \emph{International Conference on Acoustics, Speech and Signal Processing (ICASSP)}, pages 7414--7418.

\bibitem[{Sacks et~al.(1974)Sacks, Schegloff, and Jefferson}]{sacks1978simplest}
Harvey Sacks, Emanuel~A Schegloff, and Gail Jefferson. 1974.
\newblock A simplest systematics for the organization of turn taking for conversation.
\newblock \emph{Language}, 50(4):696--735.

\bibitem[{Sakuma et~al.(2023)Sakuma, Fujie, and Kobayashi}]{sakuma2023slt}
Jin Sakuma, Shinya Fujie, and Tetsunori Kobayashi. 2023.
\newblock Response timing estimation for spoken dialog systems based on syntactic completeness prediction.
\newblock In \emph{Spoken Language Technology Workshop (SLT)}, pages 369--374.

\bibitem[{Skantze(2017)}]{skantze2017sigdial}
Gabriel Skantze. 2017.
\newblock Towards a general, continuous model of turn-taking in spoken dialogue using {LSTM} recurrent neural networks.
\newblock In \emph{Annual Meeting of the Special Interest Group on Discourse and Dialogue (SIGdial)}, pages 220--230.

\bibitem[{Skantze(2021)}]{skantze2021review}
Gabriel Skantze. 2021.
\newblock Turn-taking in conversational systems and human-robot interaction: {A} review.
\newblock \emph{Computer Speech \& Language}, 67:101178.

\bibitem[{Stivers et~al.(2009)Stivers, Enfield, Brown, Englert, Hayashi, Heinemann, Hoymann, Rossano, De~Ruiter, Yoon et~al.}]{stivers2009universals}
Tanya Stivers, Nicholas~J Enfield, Penelope Brown, Christina Englert, Makoto Hayashi, Trine Heinemann, Gertie Hoymann, Federico Rossano, Jan~Peter De~Ruiter, Kyung-Eun Yoon, et~al. 2009.
\newblock Universals and cultural variation in turn-taking in conversation.
\newblock \emph{Proceedings of the National Academy of Sciences (PNAS)}, 106(26):10587--10592.

\bibitem[{Ward and Tsukahara(2000)}]{ward2000prosodic}
Nigel Ward and Wataru Tsukahara. 2000.
\newblock Prosodic features which cue back-channel responses in {English} and {Japanese}.
\newblock \emph{Journal of Pragmatics}, 32(8):1177--1207.

\bibitem[{W{\l}odarczak and Heldner(2016)}]{wlodarczak2016respiratory}
Marcin W{\l}odarczak and Mattias Heldner. 2016.
\newblock Respiratory turn-taking cues.
\newblock In \emph{INTERSPEECH}, pages 1275--1279.

\end{thebibliography}


\end{document}